\documentclass[sigconf]{acmart}

\usepackage{amsmath}
\usepackage{xfrac}
\usepackage{xcolor}
\usepackage{graphicx}
\usepackage{algorithm}
\usepackage{algorithmic}

\AtBeginDocument{%
  }

\setcopyright{acmlicensed}
\copyrightyear{2025}
\acmYear{2025}
\acmDOI{}

\acmConference[AdKDD ’25, Toronto, Canada]{AdKDD 2025}{August 04}{2025}

\acmSubmissionID{10}

\begin{document}

\title{SIDE: Semantic ID Embedding for effective learning from sequences}

\author{Dinesh Ramasamy, Shakti Kumar, Chris Cadonic}
\author{Jiaxin Yang, Sohini Roychowdhury, Esam Abdel Rhman, Srihari Reddy}
\email{{dineshr, shaktik, ccadonic, jiaxiny, sroychowdhury1, esam, sriharir}@meta.com}
\affiliation{\institution{Meta Platforms, Inc.}\country{}}

\renewcommand{\shortauthors}{Ramasamy et al.}

\begin{abstract}
    Sequence-based recommendations models are driving the state-of-the-art for industrial ad-recommendation systems. Such systems typically deal with user histories or sequence lengths
    ranging in the order of O($10^3$) to O($10^4$) events. While adding embeddings at this scale is manageable in pre-trained models, incorporating them into real-time prediction models is challenging due to both storage and inference costs. 
    To address this scaling challenge, we propose a novel approach that leverages vector quantization (VQ) to inject a compact \textit{Semantic ID (SID)} as input to the recommendation models instead of a collection of embeddings. Our method builds on recent works of SIDs by introducing three key innovations: (i) a multi-task VQ-VAE framework, called \textit{VQ fusion} that fuses multiple content embeddings and categorical predictions into a single Semantic ID; 
    (ii) a parameter-free, highly granular SID-to-embedding conversion technique, called \textit{SIDE}, that is validated with two content embedding collections, thereby eliminating the need for a large parameterized lookup table; and 
    (iii) a novel quantization method called \textit{Discrete-PCA} (DPCA) which generalizes and enhances residual quantization techniques.
    The proposed enhancements when applied to a large-scale industrial ads-recommendation system achieves $2.4\times$ improvement in normalized entropy (NE) gain and $3\times$ reduction in data footprint compared to traditional SID methods.
\end{abstract}


\maketitle

\section{Introduction} \label{sec:intro}

\begin{figure}
\centering
\includegraphics[width=2.8in, height=1.4in]{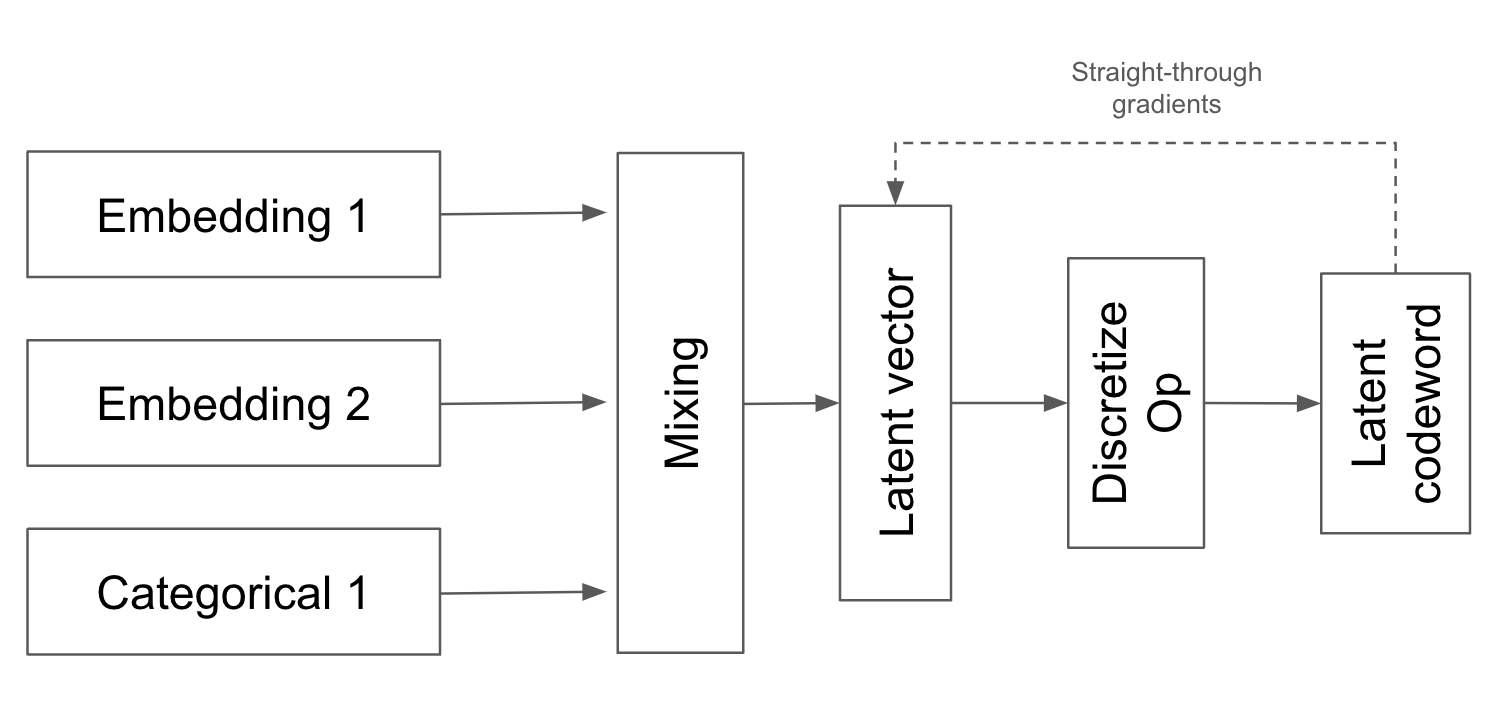}
\caption{The proposed fusion VQ-VAE setup (encoder portion). An unmixing decoder architecture (symmetric to mixing) is used to train the encoder\label{fig:fusion_ae}.}
\vspace{-0.27in}
\end{figure}

Industrial ads-ranking systems incorporate diverse signals from multiple sources to accurately predict user-engagement towards specific content. Traditionally, these systems have relied on aggregation based signals, which summarize user behavior and ad-attributes into coarse grained features. However, this approach has limitations, as it fails to capture the nuances of user interactions over time. Recent works such as in \cite{tencent2024}- \cite{alibaba2020} demonstrate the importance of long sequence user behavior for improvement in click-through-rate (CTR) prediction by leveraging O($10^3$) to O($10^4$) events. In such systems, there is a need to enrich user-engagement signals / events for improved prediction using content engagement signals, which typically are in the form of embeddings. In this work, we address two major challenges while using quantized SIDs to enhance user-ad engagement predictions: (1) How can we perform VQ efficiently when we have a plethora of content signals (embeddings and categorical predictions) as is the case in industrial ads-ranking systems, and (2) How can we structure VQ outputs to utilize them efficiently along with improved ranking performance of downstream models? To address these questions, we propose a novel multi-input vector-quantized variational auto-encoder (VQ-VAE) (called \textit{VQ fusion}), as shown in Figure~\ref{fig:fusion_ae}, which addresses the gap in existing SID approaches that require large volumes of data to relearn embeddings for different SID $n$-grams or \textit{tokens}. The proposed \textit{VQ fusion} method consists of an encoder-network that takes different content signals as input and produces a shared latent representation that is quantized and passed through a pseudo-symmetric decoder network; thereby enabling reconstruction of the different content signals. Thus, the proposed structured code-books avoid relearning the ID-to-codeword mappings and significantly reduce the number of parameters and memory required in ads-ranking.

This paper makes three major contributions. First, we propose a novel vector quantization method based on \textit{Residual Quantization} called Discrete PCA (DPCA). Second, we detail a method for fusing multiple embedding and categorical signals into a single Semantic ID which we call VQ-fusion, thus drastically reducing the data storage cost. Third, we introduce a new usage of Finite Scalar Quantization (FSQ\cite{fsq_23}) and the newly proposed DPCA methods to improve ranking outcomes using the so-called SIDE property, which stands for the ability to convert SIDs to latent embeddings in an embedding table free fashion. We demonstrate significant gains in ranking metrics (at least $2.4\times$ gain in normalized entropy) and the total return on investment (at least $7.4\times$ RoI) compared to existing approaches in an industrial ads-ranking setting.

\section{Related Work}\label{sec:rel_work}
Scalar quantization \cite{fsq_23} traditionally involves quantizing each dimension of a vector independently using a scalar codebook, which can be static (examples are \texttt{int4} and \texttt{float16} quantization which use 4 and 16 bits per dimension, respectively). A class of compression techniques involves finding a mapping to a lower dimensional subspace so that ``most'' of the information content in the embedding is preserved and the PCA method identifies the optimal linear mapping when the loss is euclidean. Vector quantization (VQ) started off with the now ubiquitous k-means clustering algorithm that can be extended in two orthogonal ways: (i) Residual k-means clustering which involves clustering the residuals to create a hierarchy of codewords and (ii) Product Quantization \cite{product_quantization_11}, which involves using parallel k-means algorithms on a subset of dimensions and is considered as a hybrid between scalar and vector quantization algorithms.

Another clustering technique is called Iterative-Quantization (ITQ, \cite{itq_13}). This involves using a linear transformation to map embeddings into the latent space where they are quantized using a binary $\{-1, 1\}$ codebook. Since the transformations are done in parallel (matrix multiplication), we classify this as a form of product quantization clubbed with auto-encoder methods. A similar auto-encoder with scalar (binary) quantization setup called Lookup-Free Quantizer (LFQ, \cite{lfq_24}) was used recently for video generation. Others have proposed Finite Scalar Quantization (FSQ, \cite{fsq_23}) which extends LFQ by allowing for more buckets per dimension. One can combine VQ based methods (e.g., k-means) with auto-encoder setup as presented in the paper on RQ-VAE\cite{rq_vae_22} that improves RQ’s ability to learn hierarchical structure by adding structured quantizer dropout, as is shown in the proposed work.

Recently there has been lot of interest in extreme scalar quantization for LLMs like 1.58-bit LLM \cite{ternary_code_llm_24}. We draw inspiration from this work and use a ternary $\{-1, 0, 1\}$ scalar codebok as a universal codebook for our proposed DPCA method. In terms of applications of VQ-techniques to ad-ranking systems like those presented in \cite{dlrm_2019}, the recent work on using RQ for defining SIDs\cite{google_rq_vae_ranking_24} and the work on generative retrieval \cite{rq_retrieval_23} stand out. Both these papers use $n$-grams of RQ codewords as SIDs and utilize these SIDs as categorical features, thereby relying largely on embedding tables. We use this approach and more traditional $k$-means as baselines in our experiments.


\section{Vector Quantization Method} \label{sec:vq}

In this section, we define our proposed method \textit{Discrete-PCA}, and expand on the techniques to combine multiple embeddings and categorical signals into one-feature via our proposed VQ-fusion method. Next, we demonstrate our embedding table-free technique called SIDE to convert codewords to embeddings in ads-ranking models. The base concept of residual quantization (RQ) \cite{google_rq_vae_ranking_24} involves the use of $D$ independent codebooks that are stacked on top of each other to compress the residue from the previous layer. The approximation to the original embedding is therefore the sum of the $D$ selected codewords. 

\subsection{Structured-Quantization method}
We propose a new quantization technique by noting that $k$-means uses a single point for each codeword. This means that the codeword collection has no inherent structure. As a result we need a large number of parameters to represent the codewords ($k$ times $d$) and we cannot increase the information content measured by $\log k$ significantly (except via product and residual quantization extensions of course). To circumvent this issue we make the codewords structured – to be precise we make groups of $L$ codewords co-linear. Thus, the codebook has the following structure:
\begin{equation}
\mathcal{C} = \Big\{~s_l \cdot \mathbf{u}_k  + \mathbf{b}_k:\forall ~k, l~\Big\} \label{eq:SRQ}
\end{equation}
where $\mathbf{u}_k$ is the codeword group’s direction vector (unit vector), $\mathbf{b}_k$ is the reference point for codeword group $k$ and $s_l$ is the signed distance of the selected codeword $l$ from the reference point $\mathbf{b}_k$ along the direction $\mathbf{u}_k$; giving us our \textit{Structured-Quantization} method.


 

Next, we present the inference logic that is leveraged by the training stage.
\subsubsection{Inference procedure} This three-step method identifies the codeword indices $k$ and $l$ given a codebook $\mathcal{C}$ and a point $\mathbf{x}$.

First, we first choose the ``line'' $\hat{k}$ which is closest to the given point $\mathbf{x}$ by measuring the distance of the point from the line corresponding the $k$-th codeword group and picking the closest line (by choosing $\mathbf{u}_{k}$):
    \begin{equation}
    \hat{k} = {\arg \min}_k \left(\left\Vert\mathbf{x} - \mathbf{b}_k\right\Vert^2 - \left|\left<\mathbf{x} - \mathbf{b}_k,\mathbf{u}_{k}\right>\right|^2\right). \label{eq:srq_distance}
    \end{equation} 
Once we identify the closest line, we estimate the optimal signed distance $\hat{s}_{\hat{k}}$ through the following inner product:
    \begin{equation}
    \hat{s}_{\hat{k}} = \left<\mathbf{x} - \mathbf{b}_{\hat{k}},\mathbf{u}_{\hat{k}}\right>.
    \end{equation}
The estimated projection weight $\hat{s}_{\hat{k}}$ is then quantized using an scalar quantizer $Q\left(\cdot\right)$ with $L$ bins to identify the codeword index $\hat{l}_{\hat{k}}$ for the signed distance. This scalar quantizer used is FSQ and is shared across the collection $\{\mathbf{u}_k\}$.

\subsubsection{Training procedure} For training this VQ-VAE model, in addition to the reconstruction loss, we add the commitment and codebook losses and use a straight-through estimator to propagate the gradients to the encoder network. 

\subsection{Towards Discrete-PCA}
In this work, we assume that a vector codebook is redundant in the context of residual and/or product quantization based structured quantization; thereby setting the the number of vector codewords to $1$. We design residual structured quantization as a form of \textit{generalized PCA} due to its stacked nature which promotes the Matryoshka property\cite{kusupati2024_matryoshka_representation_learning} of such codebooks. Since the residual depth is difficult to increase due to its serial nature, we propose to use a combination of product and residual quantization to increase the quantizer bit-width. Additionally, we design the scalar codebook ($s_{l}$ in \eqref{eq:SRQ}) by assuming that just three values $\{-1, 0, 1\}$ suffice for the signed distance. This follows recent research on Large Language Model weights, which suggest that such a ternary codebook can be universal for LLM weights \cite{ternary_code_llm_24}; thereby resulting in quantizer Discrete-PCA. The implementation of the scalar quantizer follows Finite Scalar Quantizer (FSQ) \cite{fsq_23} with number of latent dimensions equal to 1 (or the number of product quantization groups when used in conjunction with PQ).

DPCA can be viewed as a form of generalized PCA where the projection weights can only take three values $\{-1, 0, 1\}$ and thus the \textit{component vectors} no longer turn out to be orthogonal. 
The codebook collection of this method is given by the following expression:
\begin{equation}
\mathcal{C} = \sum_d \left(s_{d} \mathbf{u}_{d} + \mathbf{b}_{d}\right),\label{eq:codebook_structured_quantization}
\end{equation}
where $d$ refers to the residual depth and $s_d$ can take one of three values in $\{-1, 0, 1\}$. When used alongside product-quantization, we have $p$-parallel orthogonal codebooks, each with structure resembling \eqref{eq:codebook_structured_quantization}.

\subsection{SIDE: Converting SIDs to embeddings}\label{sub-sec:recover_latent_emb}
\begin{figure}
\centering
\includegraphics[width=\linewidth]{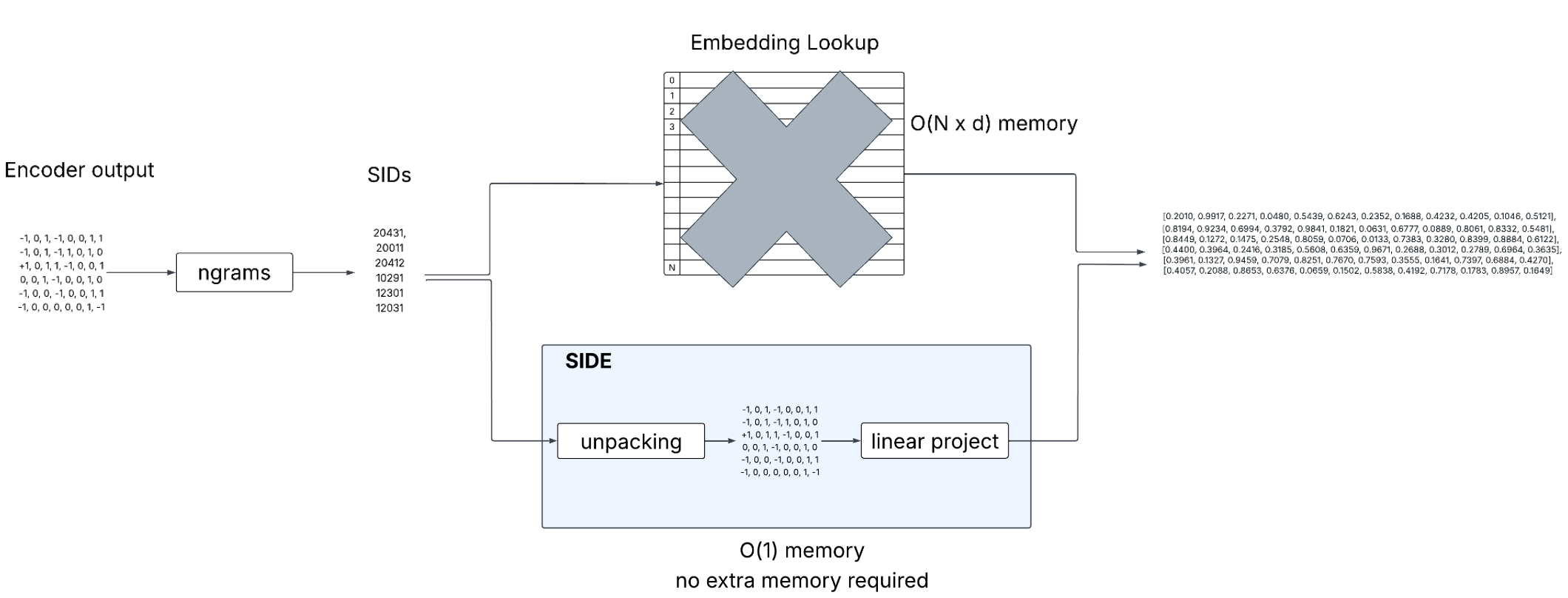}
\caption{Process of O(Nxd) memory removal using the embedding-free lookup in proposed SIDE. N = hash size of embedding tables and d=embedding dimension. The deterministic unpacking in SIDE ensures O(1) memory requirement.\label{fig:SIDE_no_emb_table}}
\vspace{-0.27in}
\end{figure}

When codewords are exported, they are typically $n$-gram-ed to form a SID or a collection of SIDs. For instance, each codeword is of the form $\mathbf{c} \in \{-1, 0, 1\}^n$ and we convert these to Semantic ID (SID) using the following $n$-gram operation:
\begin{equation}
s = \sum_{k=1}^{n} 3^k (1 + c_{k})
\end{equation}
In the ranking model, this $n$-gram operation can be undone using floor divide and modulo operations that yields the codeword vectors $\mathbf{c}$ with entries in $\{-1, 0, 1\}$. These latent codewords are called SID Embeddings or \textit{SIDE} and are used directly as embeddings. It is noteworthy that without SIDE, the memory requirement of any quantizer scales as O($\exp(C b) \times d$), where $b$ is the number of bits and $d$ is the dimension of the embedding for some constant $C$; due to embedding tables used for decoding embeddings. The proposed SIDE approach reduces this to O($b \times d$) as shown in Figure \ref{fig:SIDE_no_emb_table}.

\section{VQ Fusion} \label{sec:fusion}

In this section we describe the ads-ranking systems that utilize a multitude of content signals. These signals can be broadly categorized into two types: embeddings and categorical predictions. While incorporating all content signals into an ads-ranking model can be challenging, adding each signal one by one using SIDs obtained from VQ-methods can lead to increased storage costs and decrease the overall system efficiency. To address this challenge, we propose a novel approach of fusion, called \textit{VQ-fusion} that jointly encodes all available content signals (embeddings, categorical predictions, etc) into a single SID. Our approach leverages the multi-input multi-output auto-encoder setup,  where, the latent vector is learned using the following mixing model:
\begin{equation}
    \mathbf{h} = f\left(e_1\left(\mathbf{x}_1\right), \dots, e_n\left(\mathbf{x}_n\right)\right),
\end{equation}
from the $n$ input signals as $\{\mathbf{x}_k\}$. $e_k\left(\cdot\right)$ are the corresponding encoders for each input signal and $f\left(\cdot\right)$ is the fusion network. Next, vector quantization is performed $\mathbf{h}$ either using a version of FSQ \cite{fsq_23} (with codewords for each dimension corresponding to $\{-1, 0, 1\}$) or our proposed DPCA method in Section~\ref{sec:vq} to arrive at the codeword $\hat{\mathbf{h}}$. 
Next, $\hat{\mathbf{h}}$ (via $\mathbf{s}$ demonstrated below) is used to reconstruct the inputs $\{\mathbf{x}_k\}$ using the decoder network. This network is trained end-to-end to minimize the reconstruction loss. To propagate gradients through the non-differentiable VQ operation, $\hat{\mathbf{h}}$ is replaced with its straight-through version $
    \mathbf{s} = \mathbf{h} - \texttt{stop\_gradient}\left(\mathbf{h} - \hat{\mathbf{h}}\right)$.
    
Next, we decode the inputs from the straight-through version $\mathbf{s}$ of the latent codeword using the decoder architecture which is symmetric to the encoder as
$
\hat{\mathbf{x}}_k = g_k\left(r\left(\mathbf{s}\right)\right),
$
where $r\left(\cdot\right)$ is the shared model and $g_k\left(\cdot\right)$ are individual task decoders.
By training this joint auto-encoder to minimize weighted reconstruction loss $\sum_k w_k \ell_k\left(\mathbf{x}_k, \hat{\mathbf{x}}_k\right)$ across all content signals, a compact and informative representation can be learned that leverages the underlying relationships between the different signals. This approach leads to reduction in the cost of the semantic ID representation and improvement in the quality of representations across signals.

\section{Ads-Ranking system} \label{sec:ranking_overview}

\begin{figure}
\centering
\includegraphics[width=3in, height=2.2in]{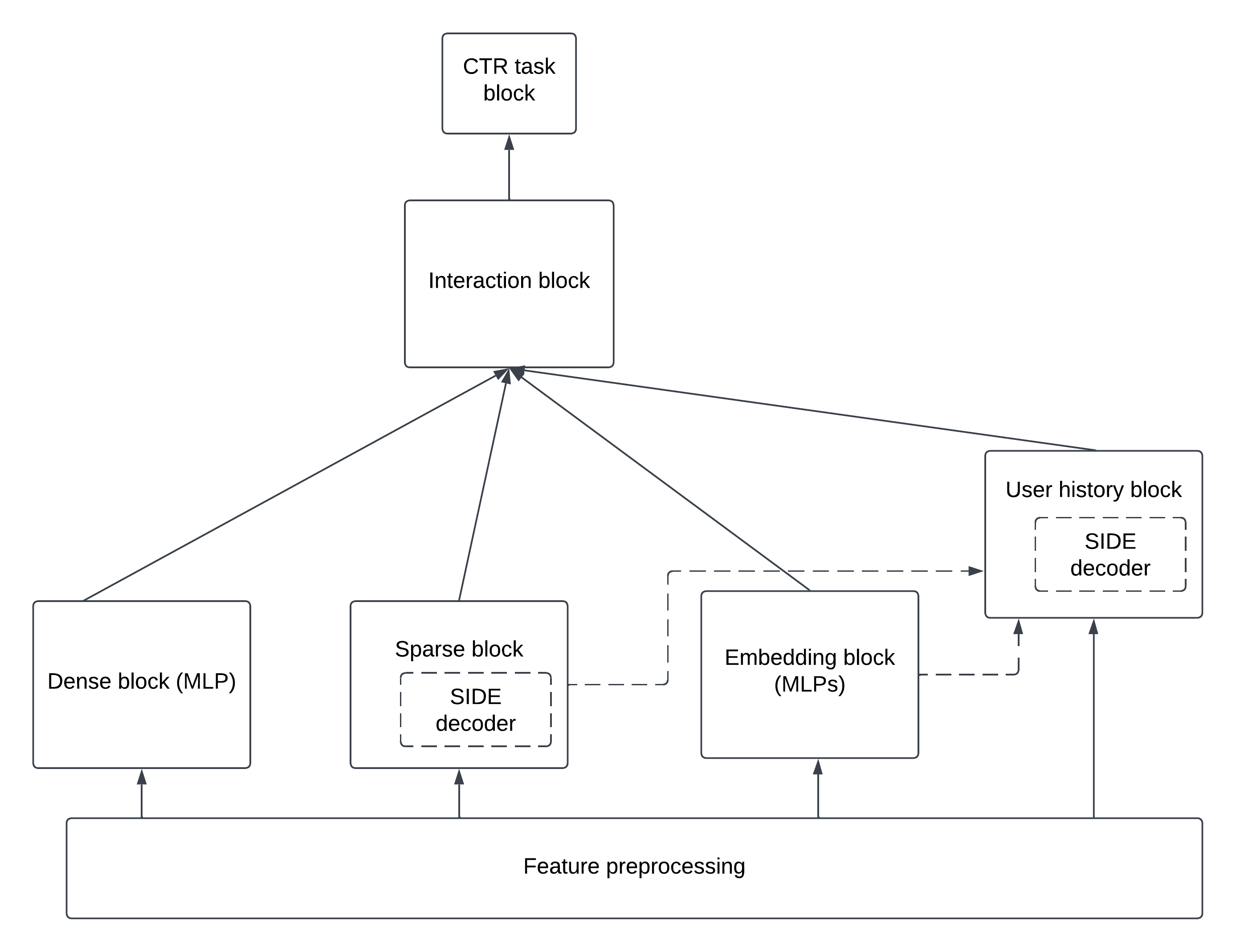}
\caption{Ranking system overview. The dotted lines represent query vectors for PMA that are typically drawn from Sparse and Embedding blocks.\label{fig:ranker_overview}}
\vspace{-0.27in}
\end{figure}
In this section we focus on the relevant components of the ads-ranking stack \cite{dlrm_2019} and highlight the changes needed to utilize the SIDE property of FSQ and DPCA based SIDs. 
Large scale deep-learning ranking systems can be broken down into (i) sparse (ii) dense (iii) embedding and (iv) sequence or user history blocks as shown in Figure~\ref{fig:ranker_overview}.

\noindent \textbf{Sparse block:} Each sparse / categorical / ID feature is mapped to a $d$-dimensional embedding $\mathbf{s}_k$ using a learnable embedding table $\mathbf{s}_k = S_k \mathbf{x}_k$,
where, $\mathbf{x}_k$ is the $r$-hot representation of the $k$-th sparse feature and $S_k$ is the associated embedding table for the feature. 

\noindent \textbf{Dense block:} All dense features are collated and represented by a single vector $\mathbf{v}$. This vector is mapped to multiple $d$-dimensional representations $\{\mathbf{d}_1, \dots, \mathbf{d}_r\}$ using learnable functions $D_k(\mathbf{v})$ for k in $\{1, \dots, r\}$. 

\noindent \textbf{Embedding block:} Embedding features are handled similar to dense features with each embedding being operated on by a separate function (typically a two-layer multi-layer perceptron) with $r=1$ outputs to yield  the $d$-dimensional representation for $k$-th embedding given by $\mathbf{e}_k$.

\noindent \textbf{User history block:} On the user-history side, the traditional approach is to convert timestamped sequence of sparse signals into embeddings using an embedding table and then leverage sequence algorithms like the Transformer-Encoder or Pooled-Multihead Attention (PMA)\cite{pma}.
The user history block is a one-layer PMA module is $
    U = \textrm{softmax}\left(\frac{Q K^T}{\sqrt{d}}\right) V$,
where, $U$ is a $k\times d$ matrix representing the $k$ user history embeddings. $Q$ is a $k \times d$ matrix derived from a subset of other sparse and embedding features $\left\{\mathbf{s}_k\right\}$ and $\left\{\mathbf{e}_k\right\}$. The key-matrix in PMA, $K = V\Theta$, where $V=\left[\mathbf{e}_1,\dots,\mathbf{e}_l\right]$ is the $l \times d$ user history embedding matrix. Here $l$ denotes the length of the user sequence and $\Theta$ is a $d \times d$ model weight matrix that maps each user history embedding to its corresponding key. We implement SIDs or their $n$-grams as sparse sequence features (based on prior work in \cite{google_rq_vae_ranking_24}) to derive the user history embeddings $\mathbf{e}_k$. 

In the following sections we demonstrate the application of the SIDE property in the user history block and compare the user-ad engagement results to that using SIDs (or their $n$-grams) directly as sparse sequence features. We convert the SIDs to embeddings $\mathbf{h}_k \in \mathbb{R}^t$ for some $t$ typically smaller than $d$ (as discussed in Section~\ref{sub-sec:recover_latent_emb}). Next, we use a $d$-dimensional projection $\mathbf{h}_k\Omega$ of these SIDE embeddings as features in the user history model, where $\Omega$ is a learnable weight matrix of shape $d \times t$.



\subsection{Online Training}
In this sub-section we present the online training components of the our VQ-fusion model (and correspondingly DPCA model) as a part of the ads-ranking system in production.
\subsubsection{Data Collection and Model Training}
The data collection process involves logging data from currently deployed upstream content models, which is then routed via logging to train our VQ-fusion encoders. The logged upstream data also serves as the ground-truth for training our auto-encoder. To ensure compliance with data privacy regulations, data is cleaned and filtered through rigorous privacy filters, such as ID masking, prior to any training. The VQ-fusion model converges quickly and needs 1 billion data points for initial offline training. Once the fusion model is trained, its inference is used to generate output features (SID) to train the downstream ranking model with ~29 billion data points for the initial offline training.

\subsubsection{Recurring Training and Inference}
Upstream content models that produce embeddings leverage recurring training through periodic scheduled training jobs. We generate pipelines for our VQ-fusion model and apply them into the existing upstream training jobs. This results in our VQ-fusion models getting trained periodically whenever upstream models are trained. This also addresses any new item embeddings which are learnt in VQ-fusion during this recurring training. The inference output (namely SID) produces IDs that are then logged into downstream training tables and serve as output features for our ads-ranking system.

\section{Experiments and Results} \label{sec:results}
In this section, we present the results for reconstruction accuracy of the SID encoding methods RQ-VAE, FSQ and our proposed DPCA in multiple settings: (i) one-embedding at a time (1:1) for two content embeddings and (ii) in the context of VQ-fusion with the same two embeddings. Also, the impact of the proposed SIDE technique in capturing dot-product ordering of the two embedding collections in the 1:1 setting is presented. Finally we analyze the overall ads-ranking performances in Section~\ref{subsec:ranking_results}.
\subsection{Encoder Design}
\subsubsection{1:1 encoder}

We implement two content embeddings from the text-only content model (for converting text → embeddings) and the image-only content model (for converting image → embeddings). For both these embeddings we use 1 billion datapoints to train the 1:1 quantizers (text embedding quantizer and image embedding quantizer using each of RQ, FSQ and DPCA). For fusion setting we jointly train a single quantizer on the combined dataset of text and image embedding using FSQ, DPCA and RQ methods. The cosine reconstruction loss of each method is shown in Table~\ref{tab:vq_isolation_fusion}.


\begin{table}[h]
    \centering
    \begin{tabular}{llcccc}
        \toprule
        \textbf{Setting} & \textbf{Method} & \textbf{Image} & \textbf{$\triangle$ \%age} & \textbf{Text} & \textbf{$\triangle$ \%age} \\
        \midrule
        1:1 & RQ & 0.1995 & - & 0.3066 & - \\
        & DPCA & 0.1870 & - & \textbf{0.2319} & - \\
        & FSQ & \textbf{0.1549} & - & 0.2435 & - \\
        \midrule
        Fusion & RQ & 0.2224 & 11.47\% & 0.2892 & 5.67\% \\
        & DPCA & \textbf{0.1945} & \textbf{4.01\%} & \textbf{0.2365} & \textbf{1.98\%} \\
        & FSQ & 0.21607 & 39.48\% & 0.2803 & 15.11\% \\
        \bottomrule
    \end{tabular}
    \caption{Cosine reconstruction loss comparison of three $24$-bit VQ methods for two $1024$-dimensional content embeddings. 1:1 refers to separate training of both the image and text quantizers while the fusion setting refers to using a single quantizer for both embeddings; thereby utilizing half the number of total bits. $\triangle$ \%age refers to the percentage increase in cosine reconstruction loss for fusion model for each method compared to 1:1 encoder model.}
    \label{tab:vq_isolation_fusion}
\vspace{-0.35in}
\end{table}
\subsubsection{Parameter tuning:} We tune the 3 major hyper-parameters as the numbers of: scalar quantization buckets, residual quantization layers, and product quantization buckets, to minimize the reconstruction loss of the embeddings. This is performed by standard parameter sweep. Once the reconstruction loss is minimized, the selected parameters are used to test loss values of the derived SIDs in ads-ranking model as sparse features to ensure loss parity or improvements in ads-ranking model performances over raw embeddings.
\subsubsection{VQ-Fusion}
We train a single quantizer on the combined dataset of image and text embeddings jointly, thereby increasing the compression by a factor of two. In Table~\ref{tab:vq_isolation_fusion}, $\Delta\%$age demonstrates the enhancement of the fusion process over 1:1 quantization method such that for image reconstruction tasks, 1:1 quantizer loss is 0.1995 and fusion quantizer loss is 0.2224 leading to $\Delta\%$age = 11.47\%.
We observe that in a fusion setting (joint training of both image and text quantizer), our proposed DPCA method has the least change in reconstruction loss when compared to FSQ and RQ methods as shown by the least $\Delta\%$age in Table~\ref{tab:vq_isolation_fusion}.

\subsubsection{Efficacy of latent codeword representation}
As discussed in Section~\ref{sub-sec:recover_latent_emb}, for FSQ and DPCA, the $n$-gram operation is reversed and centered (by subtracting $1$ in this case) to convert the Semantic IDs into embeddings. Next, the following steps are performed for the 1:1 encoder: The kNN results (using cosine similarity) are computed for the 1:1 $24$-bit compression of the two $1024$-dimensional content embeddings presented in Table~\ref{tab:semantic_to_latent}. These kNN results are then compared to those obtained from the original raw embeddings' closest $20$ kNN results (based on cosine similarity as well). We use Recall@$k$ for different values of $k$ to quantify the correctness of this comparison by using $1000$ random seed queries picked from a $200$K large embedding corpus. 
\begin{table}[ht]
    \centering
    \begin{tabular}{llccc}
        \toprule
        \textbf{Corpus} & \textbf{Method} & \textbf{R@20} & \textbf{R@50} & \textbf{R@100} \\
        \midrule
        Image & DPCA & 0.1467 & 0.3326 & 0.5088 \\
         & FSQ & \textbf{0.2370} & \textbf{0.4187} & \textbf{0.5998} \\
        \midrule
        Text & DPCA & \textbf{0.1703} & \textbf{0.3418} & \textbf{0.5213} \\
         & FSQ & 0.1323 & 0.2943 & 0.4848 \\
        \bottomrule
    \end{tabular}
    \caption{Comparison of DPCA and FSQ methods using Recall at 20, 50 and 100 for closest-20 neighbors as defined by the original embedding to show the efficacy of Semantic ID to latent embedding conversion (SIDE property).}
    \label{tab:semantic_to_latent}
    \vspace{-0.25in}
\end{table}

\begin{figure}[h]
  \centering
  \includegraphics[width=3in, height=1.3in]{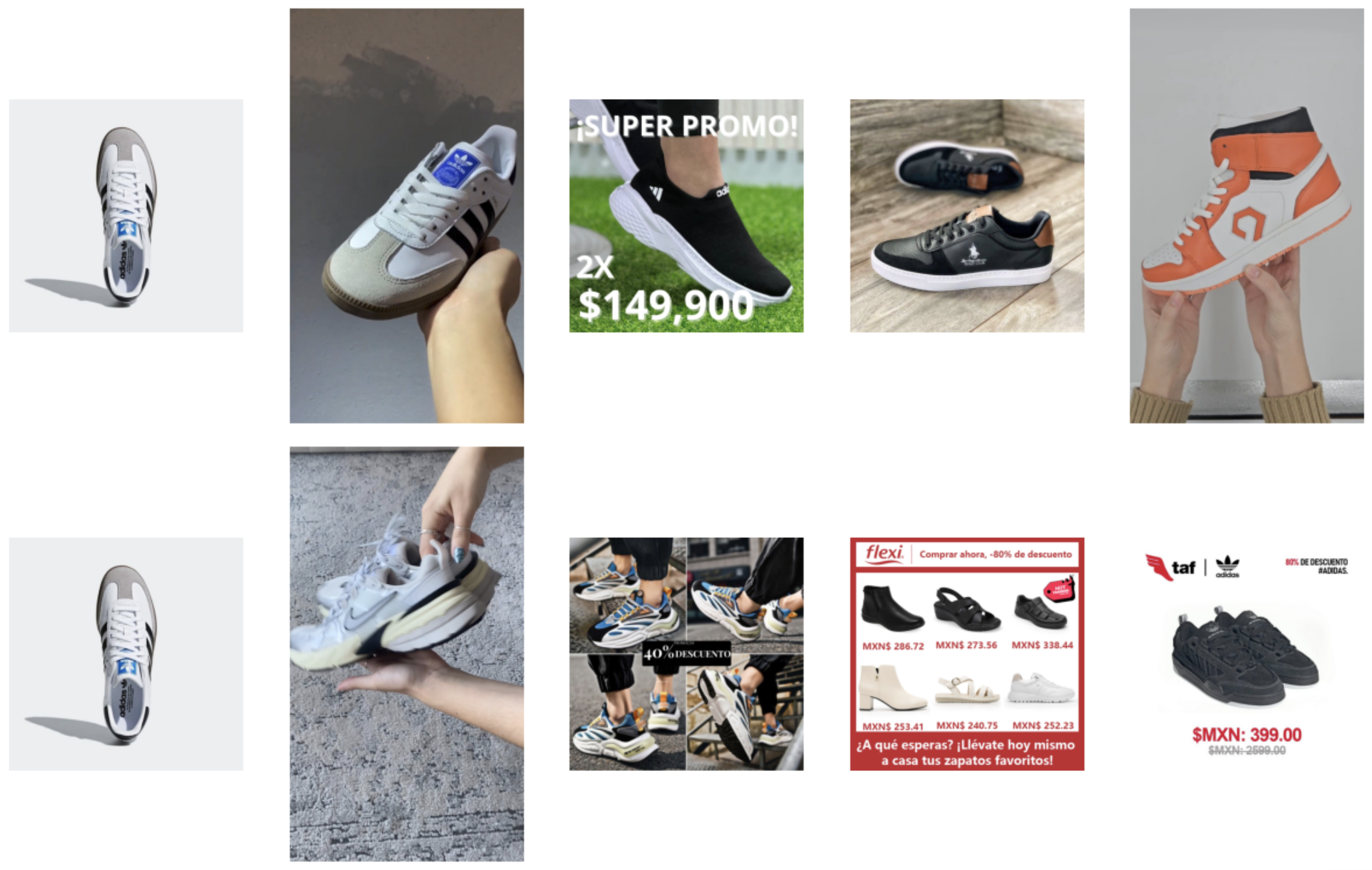}
  \caption{k-NN results with leftmost ad as the query using cosine similarity as the measure. Top: Thumbnails corresponding to original image embedding. Bottom: Thumbnails corresponding to 680x compressed FSQ SIDE representation. All of these ads are related to sneakers.}
  \label{fig:knn_results_image_fsq}
  \vspace{-0.27in}
\end{figure}

Based on Table~\ref{tab:semantic_to_latent}, we observe that both FSQ and proposed DPCA retain the dot-product ordering information present in the original embedding even with a massive $1024 \times 16 \div 24 \approx 682$ compression ratio. We verify these results qualitatively by inspecting k-NN results like those in Figure~\ref{fig:knn_results_image_fsq}.

\subsection{Ads-Ranking}\label{subsec:ranking_results}
In our ads-ranking systems, we predict two main tasks: click (CTR) and the conversion (CVR: eventual purchase) probabilities. In this paper we access the improvement for CTR prediction in terms of Normalized Entropy, which is defined as follows:
\begin{equation}
    \textrm{NE} = \frac{\sfrac{1}{N}\sum \left(y_i \log p_i + (1- y_i) \log (1-p_i)\right)}{\hat{p} \log\hat{p} + \left(1 - \hat{p}\right)\log\left(1 - \hat{p}\right)},
\end{equation}
where, $y_i$ are the labels, $p_i$ are model predictions, $\hat{p} = \sfrac{\sum y_i}{N}$ is the prior probability and $N$ is the number of samples. We utilize $\frac{\left(NE_{\textrm{click}} + NE_{\textrm{conv}}\right)}{2C}$ to compute the overall RoI of the feature, where $C$ is the logging cost of onboarding the feature.

In this experiment, we analyze the impact of adding K-means top-$k$ clusters ($k$ nearest neighbors) and SIDs as sparse-inputs (corresponding to the ad-item to be ranked) in Table~\ref{tab:ranking_results_ad_side} for a new content feature. Note that we use the same hash size for both K-means and SIDs. We now demonstrate the NE gains and RoI for a DPCA encoded feature used as an ad-feature in our ranking architecture. 
\begin{table}[h]
    \vspace{-0.15in}
    \centering
    \begin{tabular}{lccc}
        \toprule
        \textbf{Method} & $k$-means & SID\\
        \midrule
        \textbf{Incremental Cost} & $\times 1$ & $\times 0.23$ \\
        \midrule
        \textbf{Click NE gain} & \textbf{0.0108\%} & 0.0085\% \\
        \midrule
        \textbf{Conversion NE gain} & 0.0037\% & \textbf{0.0101\%} \\
        \midrule
        \textbf{RoI} & $\times$1 & $\times 5.57$ \\
        \bottomrule
    \end{tabular}
    \caption{Incremental data and feature cost and normalized entropy gain relative to the baseline for a DPCA encoded ad feature. Baseline is a large-scale production ads-ranking model without the specific ad-features.}
    \label{tab:ranking_results_ad_side}
    \vspace{-0.25in}
\end{table}

In Table~\ref{tab:ranking_results_ad_side}, the cost $C$ of features is directly proportional to the feature length used. $k$-means features with length $50$ have cost $46.54kW$, while the SID features have length $3$ and corresponding cost is $10.84kW$. This leads to a reduction in incremental feature cost by $76.7\%$. Hence from Table~\ref{tab:ranking_results_ad_side} we see that SID improves RoI by $5.57$ times over traditional $k$-means method. This leads to an NE gain for using DPCA as 1.28X or 28\%

Next, we present results for adding SID sparse inputs and SIDs recovered as latent embeddings (SIDE) for the user-history block in two settings: (i) 1:1 encoder (in Table~\ref{tab:ranking_results_user_side}) and VQ-fusion (in Table~\ref{tab:ranking_results_user_side_fusion}) for two different types of user history sequences. 
\begin{table}[h]
    \centering
    \begin{tabular}{lccc}
        \toprule
        \textbf{Method} & SID & SIDE\\
        \midrule
        \textbf{Incremental cost} & $\times 1$ & $\times 0.33$ \\
        \midrule
        \textbf{Click NE gain} & 0.0082\% & \textbf{0.0185\%} \\
        \midrule
        \textbf{Conversion NE gain} & 0.0086\% & \textbf{0.0111\%} \\
        \midrule
        \textbf{RoI} & $\times1$ & \textbf{$\times 5.33$} \\
        \bottomrule
    \end{tabular}
    \caption{Normalized entropy gain relative to the baseline for a DPCA encoded feature used as user history. Baseline is the large-scale production ads-ranking model.}
    \label{tab:ranking_results_user_side}
    \vspace{-0.27in}
\end{table}
In Table~\ref{tab:ranking_results_user_side}, we demonstrate the NE gains for using SID and unpacking it with SIDE for a DPCA encoded feature. The feature cost for SIDE is 1/3rd of SID since we just use a single $n$-gram from SID to unpack in SIDE. We observe that SIDE further improves NE gains over SID and enhances the RoI by $5.33\times$ for the DPCA encoded feature on the user-side. Note that the hash size in SID is set to number of scalar quantization raised to power $n$-gram length, which is $64^3=$ max cardinality of the SID feature, as shown in Section \ref{section-hparam-tuning}.

In Table~\ref{tab:ranking_results_user_side_fusion}, the gains from SIDE are further enhanced when by using a VQ-fusion SID feature comprising of $6$ individual signals encoded into a single feature as. Here, the RoI improves by $7.40\times$ over SID. This leads to an NE gain when using SIDE to be 2.44x or 144\%.
\begin{table}[h]
    \vspace{-0.15in}
    \centering
    \begin{tabular}{lccc}
        \toprule
        \textbf{Method} & SID & SIDE\\
        \midrule
        \textbf{Incremental cost} & $\times 1$ & $\times 0.33$ \\
        \midrule
        \textbf{Click NE gain} & 0.0100\% & \textbf{0.0284\%} \\
        \midrule
        \textbf{Conversion NE gain} & 0.0067\% & \textbf{0.0124\%} \\
        \midrule
        \textbf{RoI} & $\times$1 & \textbf{$\times 7.40$} \\
        \bottomrule
    \end{tabular}
    \caption{Normalized entropy gain for VQ fusion encoded feature comprising $6$ individual signals, used as user history. Baseline = 6 different 1:1 encoded DPCA features used as SID.}
    \vspace{-0.35in}
    \label{tab:ranking_results_user_side_fusion}
\end{table}

\subsubsection{Encoder hyper-parameterization}\label{section-hparam-tuning}
From our ads-ranking experiments we demonstrate examples of varying the SID construction hyper-parameters and its impact in NE gains from the ads-ranking model.
In the VQ-Fusion experiment, summarizing 6 user-side features in Table \ref{tab:ranking_results_user_side_fusion} and retaining the number of scalar quantization buckets as 64 and the number of residual quantization layers as 9 and increasing the $n$-gram size in SID, leads to the NE trend in Table \ref{tab:ablations}. 
\begin{table}[h]
    \vspace{-0.10in}
    \centering
    \begin{tabular}{lccc}
        \toprule
        \textbf{Prefix n-gram length} & SID & SIDE\\
        \midrule
        3 & 0.029\% & 0.044\% \\
        \midrule
        4 & 0.020\% & 0.046\% \\
        \bottomrule
    \end{tabular}
    \caption{Effect of prefix $n$-gram length on NE, keeping all other Encoder hyper-parameters fixed.}
    \label{tab:ablations}
    \vspace{-0.25in}
\end{table}
Here, we observe that for SID, as more $n$-gram IDs are introduced (by making the prefix $n$-gram longer from 3 to 4) in the ads-ranking model, the NE gains reduce due to higher noise due to increase in hash collisions based on the embedding table size limitations. The maximum hash size for $n$-gram length 3 is $64^3=262,144$ and for length 4 is $64^4=16,777,216$.
However, our proposed SIDE technique avoids hash collisions while  unpacking the 4th $n$-gram ID leading to NE increase.

We also assess the effect of increasing the number of scalar quantizer codewords on the ads-ranking NE gains. Using 4 scalar quantizer codewords NE gain for SID and SIDE are 0.018\% and 0.035\% respectively. Extending this analysis to 64 codewords results in NE gains for SID and SIDE as 0.015\% and 0.044\%, respectively.

\subsection{Discussion on Production Impact}
In this section we assess the end-user benefits in terms of ads-score (values for both advertisers and users). Ads-score is the sum of the two entities: 1) ads-value or the value to advertisers calculated by adding paced-bids, 2) quality value which measures value to users, considering,
positive interactions (e.g., clicks, likes), negative interactions (e.g., hiding an ad), ad-quality and user engagement. We observe overall gains of ~0.17\% ad-score with the launch bundle involving our proposed method (DPCA, VQ-fusion and SIDE put together) on the end-users. 

In terms of other production metrics: our VQ-Fusion models deployed in online inference have very low latency (< 500 ms) and memory consumption (<= 2.5 GB), recurring training has ~0.33\% QPS regression. The inference model requires 2, 16GB T1 machines, leading to 20$\times$ lesser deployment machine costs when compared to traditional quantization models like k-means leading to improved RoI for our proposed system. The latency and throughput effect on feature logging due to the additional VQ-Fusion model is negligible, since real-time feature availability for ads-ranking systems remains less than 60 minute SLA-bounds.

\section{Conclusion and Future Work}\label{sec:conclusion}

In this paper, we have presented a novel approach to incorporating high-dimensional content embeddings into sequence-based recommendation models using vector quantization. Our method called VQ-fusion, which leverages a multi-task VQ-VAE fusion framework and an embedding table-free SID to embedding conversion technique (SIDE), has been shown to reduce storage costs while improving ranking outcomes on large-scale ads recommendation systems.  
The SIDE method can lead to significant improvements in the efficiency for the sequence learning paradigm in modern ad-recommendation systems.


\section{Acknowledgements}
The authors would like to thank our collaboration partners in meeting our deployment goals. For Deployment and Launch teams: Doris Wang, Parshva Doshi, Mingwei Tang, Xinlong Liu, Leo Ding, Ahmed Agiza, Rahul Mayuranath, Dave Li. For Quantization support: Shuby Mao, Jijie Wei, Michael Chang. For Preproc support: Varun Bhadauria, Priya Ramani, Kapil Sharma, Tao Jia Sahaj Biyani. Package release: Pablo Ruiz, Fischer Bennetts, Hang Qi, Zhongjie Ma, Ying Xu. Infra and testing: Abdul Zainul-Abedin, Zhaoyang Huang, Liang Tao. Management support: Kapil Gupta, Rostam Shirani, Arnold Overwijk.

\bibliographystyle{ACM-Reference-Format}
\bibliography{references}


\end{document}